\documentclass[natbib, a4paper]{svproc}

\usepackage[numbers]{natbib}
\makeatletter 
\renewcommand\@biblabel[1]{#1.} 
\makeatother

\usepackage[utf8]{inputenc}

\usepackage[bookmarks=true]{hyperref}

\hypersetup{
    breaklinks=true,
    colorlinks=true,
    hidelinks
}

\usepackage{xcolor}
\definecolor{espblack}{RGB}{0,0,0}
\definecolor{espwhite}{RGB}{255,255,255}
\definecolor{espgray}{RGB}{206,206,206}
\definecolor{esplightgray}{RGB}{224,224,224}
\definecolor{espdarkgray}{RGB}{168,168,168}
\definecolor{espsomewhatdarkgray}{RGB}{130,130,130}
\definecolor{espverydarkgray}{RGB}{100,100,100}
\definecolor{espblue}{RGB}{11,93,174}
\definecolor{esplightblue}{RGB}{59,175,236}
\definecolor{espdarkblue}{RGB}{6,26,64}
\definecolor{espred}{RGB}{206,62,21}
\definecolor{esplightred}{RGB}{206,62,21}
\definecolor{espdarkred}{RGB}{61,19,8}
\definecolor{espyellow}{RGB}{232,163,26}
\definecolor{espgreen}{RGB}{100,161,27}
\definecolor{esplightgreen}{RGB}{149,198,35}
\definecolor{espdarkgreen}{RGB}{49,92,43}
\definecolor{esppurple}{RGB}{106,20,125}
\definecolor{esplightpurple}{RGB}{197,137,232}
\definecolor{espdarkpurple}{RGB}{50,14,59}
\usepackage{tikz}

\usepackage{tikzscale}

\usepackage{pgfplots}

\usepackage{pgfplotstable}

\usetikzlibrary{calc}

\usetikzlibrary{positioning}

\usepgfplotslibrary{statistics}

\usepgfplotslibrary{fillbetween}

\pgfplotsset{
  compat=1.15,
  mps basic/.style={
    xlabel near ticks,
    xlabel style={font=\footnotesize},
    ylabel near ticks,
    ylabel style={font=\tiny},
    xmajorgrids,
    major x grid style={dotted},
    ymajorgrids,
    major y grid style={dotted},
    tick label style={font=\tiny}
  },
  mps scientific x/.style={
    x tick label style={
      /pgf/number format/sci,
      font=\tiny
    }
  },
  mps scientific y/.style={
    y tick label style={
      /pgf/number format/sci,
      font=\tiny
    }
  },
  mps fixed x/.style={
    x tick label style={
      /pgf/number format/.cd,
      fixed,
      fixed zerofill,
      precision=6,
      /tikz/.cd,
      font=\tiny
    }
  },
  mps fixed y/.style={
    y tick label style={
      /pgf/number format/.cd,
      fixed,
      fixed zerofill,
      precision=6,
      /tikz/.cd,
      font=\tiny
    }
  },
  /pgfplots/myylabel absolute/.style={%
      /pgfplots/every axis y label/.style={at={(0,0.5)},xshift=#1,rotate=90,align=center},
      /pgfplots/every y tick scale label/.style={
        at={(0,1)},above right,inner sep=0pt,yshift=0.3em
      }
   }
}

\usetikzlibrary{decorations.pathreplacing,shapes.misc}
\tikzset{
  cross/.style={cross out, draw=black, minimum size=2*(#1-\pgflinewidth), inner sep=0pt, outer sep=0pt},
  cross/.default={4pt},
  start/.style={draw = black, fill = black, circle, inner sep = 0pt, minimum size=3pt},
  goal/.style={draw = black, fill = white, circle, inner sep = 0pt, minimum size=3pt},
  vertex/.style={draw = black, fill = black, circle, inner sep = 0pt, minimum size=0.4pt},
  rgg edge/.style={densely dotted, very thin, espblack},
  forward edge/.style={espblack, thin},
  reverse edge/.style={espgray, thin},
  invalid edge/.style={espblack, thick, densely dotted},
  obstacle/.style={fill = espblack, draw = espblack},
  antiobstacle/.style={draw = none, fill = white},
  boundary/.style={draw = black, fill = none},
  start_region/.style={draw = esplightgreen, ultra thick, fill = none},
  goal_region/.style={draw = espred, ultra thick, fill = none},
  solution/.style={espyellow, very thick}
}

\usepackage[linesnumbered,vlined,ruled,commentsnumbered]{algorithm2e}
\SetKwRepeat{Do}{do}{while}

\SetVlineSkip{0.2em}

\SetInd{0.1em}{0.2em}

\DontPrintSemicolon{}

\SetKwIF{If}{ElseIf}{Else}{if}{}{else if}{else}{endif}

\SetAlgoSkip{}

\SetAlCapHSkip{0cm}

\SetAlgoCaptionLayout{footnotesize}

\SetAlCapNameFnt{\scriptsize}
\SetAlCapFnt{\scriptsize}

\SetArgSty{textnormal}

\makeatletter
\patchcmd\algocf@Vline{\vrule}{\vrule \kern-0.4pt}{}{}
\patchcmd\algocf@Vsline{\vrule}{\vrule \kern-0.4pt}{}{}
\makeatother

\makeatletter
\patchcmd{\@algocf@start}%
  {-1.5em}%
  {0pt}%
  {}{}%
\makeatother

\usepackage{etoolbox}

\robustify\bfseries
\newrobustcmd\B{\DeclareFontSeriesDefault[rm]{bf}{b}\bfseries}  

\usepackage{tabularx}
\usepackage{booktabs}
\usepackage{siunitx}

\usepackage{times}
\usepackage{helvet}
\usepackage{courier}

\usepackage{mathtools}
\usepackage{amsfonts}

\usepackage[bottom]{footmisc}

\usepackage{microtype}

\usepackage{graphicx}
\usepackage[font={small}]{caption}
\usepackage[font={small}]{subcaption}

\usepackage[capitalise]{cleveref}
\crefname{algorithm}{Alg.}{Algs.}

\usepackage{blindtext}

\usepackage{stackengine}
\setstackgap{S}{2pt}

\usepackage{pgfplots}
\usepgfplotslibrary{fillbetween}

\pgfplotsset{compat=1.15}

\usepackage{float}

\newcommand{\effort}{\bar{e}}
\newcommand{\apeffort}{\bar{d}}

\newcommand{\EI}{EIRM*}
\DeclareMathOperator*{\argmin}{arg\,min}    %

\newcommand{\setInsert}[0]{\!\xleftarrow{\scriptscriptstyle +}\!}
\newcommand{\setRemove}[0]{\!\xleftarrow{\scriptscriptstyle -}\!}

\begin{document}
\mainmatter

\title{Effort Informed Roadmaps (\EI{}): Efficient Asymptotically Optimal Multiquery Planning by Actively Reusing Validation Effort}

\titlerunning{Effort Informed Roadmaps (\EI{})}

\author{Valentin N. Hartmann\inst{1}\and Marlin P. Strub\inst{2}\and Marc Toussaint\inst{1}\and Jonathan D. Gammell\inst{3}%
}
\authorrunning{Valentin N. Hartmann et al.}

\institute{Learning and Intelligent Systems Group, TU Berlin, Germany
\and
Jet Propulsion Laboratory, California Institute of Technology, USA. Work
performed while at the University of Oxford.
\and
Estimation, Search, and Planning (ESP) Group of the Oxford Robotics Institute (ORI), University of Oxford, United Kingdom}

\maketitle
\setcounter{footnote}{0}

\begin{abstract}

Multiquery planning algorithms find paths between various different starts and goals in a single search space.
They are designed to do so efficiently by reusing information across planning queries.
This information may be computed before or during the search and often includes knowledge of valid paths.

Using known valid paths to solve an individual planning query takes less computational effort than finding a completely new solution.
This allows multiquery algorithms, such as PRM*, to outperform single-query algorithms, such as RRT*, on many problems but their relative performance depends on how much information is reused.
Despite this, few multiquery planners explicitly seek to maximize path reuse and, as a result, many do not consistently outperform single-query alternatives.

This paper presents Effort Informed Roadmaps (\EI{}), an almost-surely asymptotically optimal multiquery planning algorithm that explicitly prioritizes reusing computational effort.
\EI{} uses an asymmetric bidirectional search to identify existing paths that may help solve an individual planning query and then uses this information to order its search and reduce computational effort.
This allows it to find initial solutions up to an order-of-magnitude faster than state-of-the-art planning algorithms on the tested abstract and robotic multiquery planning problems.\looseness=-1

\keywords{sampling-based path planning, optimal path planning, multiquery path planning}

\end{abstract}

\section{Introduction}
A general-purpose path planner aims to find a path that connects a start to a goal, typically in a continuous space.
The underlying structure of many environments is static and tends to pose repetitive problems, such as in home \cite{faust2018prmrl}, construction \cite{funk2021learn2assemble,hartmann2020robust}, or kitchen \cite{lagriffoul2018platform} scenarios. %
Multiquery planners are designed to solve multiple different start-goal queries in  static environments by exploiting this repetitiveness to reduce the computational time required to find a solution.
A large component of this computational effort for an individual planning query is checking if a path is collision free, i.e., validation effort~\cite{hauser2015lazy,Solovey2020,sanchez2002}.
This can be reduced in multiquery settings by reusing previously gained knowledge of valid edges to solve subsequent queries more efficiently.

Many planners compute reusable information in advance.
Probabilistic Roadmaps (PRM)~\cite{kavraki1996probabilistic} construct a roadmap and collision check all its edges during preprocessing.
This roadmap is then used to simplify individual queries to a graph search over the roadmap, resulting in fast solution times.
\begin{figure}[t]
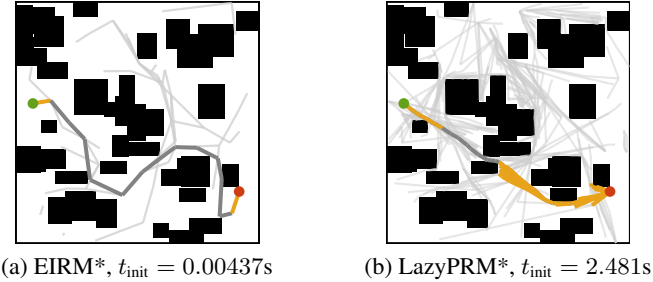

\centering
    \begin{subfigure}[t]{.4\textwidth}
        \centering
        \input{img/intro/eiprm}
        \caption{\EI{}, $t_\text{init} = 0.00437\text{s}$ }
    \end{subfigure}%
    \begin{subfigure}[t]{.4\textwidth}
        \centering
        \input{img/intro/lprm}
        \caption{LazyPRM*, $t_\text{init} = 2.481\text{s}$}
    \end{subfigure}%
    
    \caption{An illustration of \EI{} finding the initial solution orders-of-magnitude faster than LazyPRM* for the 20$^\text{th}$ query of a multiquery problem.
    The green and red disks are the start and the goal, respectively.
    Edges that are collision checked in this query are yellow, and edges that are reused are dark grey.
    Light grey edges are valid edges that have been collision checked in previous queries. The time of the initial solution is $t_\text{init}$.
    \EI{} validates two edges to connect to the preexisting graph while LazyPRM* validates 81 edges.}
    \label{fig:pull}
\end{figure}

If the environment is not available in advance, reusable information needs to be calculated in parallel to solving queries.
LazyPRM \cite{bohlin00lazy} solves individual queries by optimistically assuming all edges and vertices in the roadmap are valid, and then only checking the edges and vertices of the solution that are not yet validated.
It preserves the knowledge of validated and invalidated edges over multiple queries but does not \textit{actively} reuse previously invested effort, so any reuse of effort is by coincidence.

PRM*~\cite{karaman2011sampling} and LazyPRM*~\cite{hauser2015lazy} extend these ideas to the optimal planning problem by continually adding more samples to their roadmaps to improve the approximation with additional computational time. 
This roadmap asymptotically contains the optimal solution with probability one, i.e., is almost-surely asymptotically optimal.
While this graph growth improves solution quality, it can also increase initial solution times since new queries start from the previous approximation and searching large graphs can be prohibitively expensive.

This paper presents Effort Informed Roadmaps (\EI{}), an almost-surely asymptotically optimal anytime multiquery planner which seeks to to prioritize finding an initial solution by actively reusing effort from previous queries.
\EI{} extends Effort Informed Trees (EIT*)~\cite{strub2021ait} to the challenges of solving multiquery problems quickly by:
\textit{(i)} actively seeking to reuse computational effort, and \textit{(ii)} managing graph size over multiple planning queries.

\EI{} quickly finds an initial solution to a planning query by using a search explicitly informed by validation effort.
It then uses a cost-informed search to improve this solution by efficiently adding and searching more samples for as long as time allows for the current query.
When a new planning query is posed, EIRM* prevents the complexity of this high-resolution graph from negatively affecting search performance by \emph{rewinding} the approximation to the first batch of samples.
It then reuses computational effort to solve this new query by both informing the initial search by validation effort and improving the solution by \emph{replaying} the previous samples and reusing any previously validated edges.
This allows EIRM* to find initial solutions to individual queries faster than other planners while almost-surely converging asymptotically to the same global optimum.\looseness=-1

We compare \EI{} to other planners available in Open Motion Planning Library (OMPL)~\cite{sucan2012ompl} on several low-{} and high-dimensional abstract environments, and simulated robots.
On these problems, it solves later queries of a problem up to an order-of-magnitude faster than the tested planners while performing the same on initial queries.

\section{Related Work}
A general overview of sampling-based motion planning can be found in \cite{elbanhawi2014sampling, gammell2021asymptotically} and a review of work using search effort, including the single-query Bayesian Effort-Aided Search Trees (BEAST) \cite{kiesel2017}, can be found in \cite{strub2021ait}.
This review focuses on multiquery planning, which aims to enable efficient planning over multiple queries by reusing knowledge gained from previous queries in the planning for the current query.

Many multiquery algorithms are based on Probabilistic Roadmaps (PRM) \cite{kavraki1996probabilistic}.
PRM randomly samples states in the configuration space of the robot to build a roadmap that discretizes the space.
It validates all edges in the roadmap in a preprocessing phase before solving the first query.
A query is then solved by a search over the roadmap.

LazyPRM \cite{bohlin00lazy} avoids the preprocessing and is more suitable for problems when the environment is not known in advance.
It initially assumes all edges and vertices are valid, and searches over the graph to find a potential solution.
It then collision checks the edges and vertices of this solution candidate that have not been collision checked in previous queries.
This approach does not \emph{actively} exploit the knowledge of collision checked edges and any reduction in planning effort is not deliberate.

PRM*~\cite{karaman2011sampling} and LazyPRM*~\cite{hauser2015lazy} extend PRM and LazyPRM to obtain almost-sure asymptotic optimality by adding new samples to the roadmap once a solution is found to improve the approximation of the environment.
This allows them to probabilistically converge towards the global optimum but, together with the starts and goals from previous queries, these added samples increase the size of the roadmap in every query, and slow down subsequent queries.

Sparse Roadmap Spanners (SPARS) \cite{dobson2013sparse} address the unbounded growth of the graph by storing previously found paths in a sparse graph.
The construction of the sparse graph is computationally expensive, and can lead to a slower planner overall.
SPARS2~\cite{dobson2014sparse} improves the construction of the graph and reduces memory requirements, but it is still computationally expensive. %

Experience Graphs (E-Graph) \cite{phillips2012graphs} contain previously found solutions and enable combining parts of these paths to find solutions to new queries.
E-Graphs require running an all-pairs shortest path algorithm, which is computationally expensive.
E-Graphs are extended to handle incremental anytime planning by continually updating cost heuristics \cite{phillips2013anytime}.
This is also computationally expensive and does not always improve total planning time.
The increasing graph size is not considered, which may lead to unsustainable graph sizes for high number of queries.

Lightning \cite{berenson2012robot} and Thunder \cite{coleman2015experience} are planning frameworks that store previous solutions in a database and attempt to modify these solutions to solve the current path planning query.
The database in Lightning can grow unbounded, which leads to slow information retrieval.
Thunder accelerates information retrieval by storing solutions to previous planning problems in a sparse graph, which eliminates redundant information. 
As in SPARS, the complexity of inserting paths in the sparse graph can be computationally expensive.
Lightning and Thunder only store and reuse solutions and do not retain otherwise validated or invalidated edges between queries.

A different way to avoid the graph growth is to use single-query path planners and modify them to reuse results from previous queries.
\citet{bruce2002real} bias Rapidly Exploring Random Trees (RRTs) \cite{Lavalle98rapidly} with samples from previously found paths that are stored in a waypoint cache.
This cache has a fixed size and old samples are eventually forgotten, which requires problem-specific tuning.
Other approaches to speed up the planning process learn to sample specific features (e.g., narrow passages) efficiently \cite{chen2019learning, ichter2018learning, ichter2020learned}. 
This does not actively reuse planning effort, and may require preprocessing to learn the sampling distribution. %

Reconfigurable Random Forests (RRF) \cite{li2002incremental} extend RRTs to the multiquery setting.
RRF grows trees anchored at the start and the goal of a query towards the previously constructed trees, in order to iteratively construct a roadmap.
This approach does not result in an anytime planner and does not give any optimality guarantees.

In comparison to the planners reviewed in this section, \EI{} actively tries to find initial solutions quickly.
\EI{} estimates remaining validation effort to inform its initial search. 
This allows \EI{} to actively reuse previous computational effort to find initial solutions quickly, contrary to LazyPRM*.
Calculating this effort heuristic is computationally inexpensive compared to the approach taken by E-Graphs.
\EI{} prevents the work from earlier queries slowing down later queries by actively managing graph size.
Unlike SPARS, \EI{} does this efficiently by rewinding the graph of each query to the initial approximation.

These approaches allow \EI{} to find initial solutions quickly and almost-surely converge asymptotically to the global optimum of each planning query.

\begin{figure*}[t]
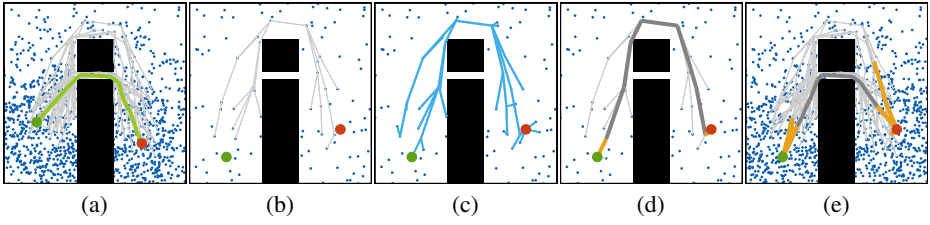

\captionsetup[subfigure]{aboveskip=3pt}
\captionsetup[subfigure]{belowskip=-2pt}
\centering
    \begin{subfigure}[t]{.2\textwidth}
        \centering
        \input{img/eiprm/illustration/q_19}
        \caption{}
    \end{subfigure}\hfill%
    \begin{subfigure}[t]{.2\textwidth}
        \centering
        \input{img/eiprm/illustration/initial}
        \caption{}
    \end{subfigure}\hfill%
    \begin{subfigure}[t]{.2\textwidth}
        \centering
        \input{img/eiprm/illustration/rev_tree}
        \caption{}
    \end{subfigure}%
    \begin{subfigure}[t]{.2\textwidth}
        \centering
        \input{img/eiprm/illustration/effort_sol}
        \caption{}
    \end{subfigure}\hfill%
    \begin{subfigure}[t]{.2\textwidth}
        \centering
        \input{img/eiprm/illustration/final_sol}
        \caption{}
    \end{subfigure}%
    \caption{\label{fig:illustration} An illustration of \EI{} solving the 20$^\text{th}$ query of a multiquery problem.
    The states are blue dots, the start is the green disk, the goal is the red disk, previously validated edges are light grey, newly validated edges in this iteration are yellow, the reverse tree is light blue, and edges that are reused in the solution are dark grey. 
    The solution to the 19$^\text{th}$ planning query is shown in green along with the available previously validated edges at this resolution, (a).
    The 20$^\text{th}$ query is solved by first resetting the approximation to the initial batch and its validated edges, (b).
    The reverse tree is then grown to the start, (c). 
    Since the reverse search is initially ordered purely on effort, the previously validated edges (which have zero validation effort) are explored first. %
    A solution is found in the current approximation by only validatating two new edges, (d).
    The approximation is continually refined and the search is ordered by cost to find better solutions, (e).
	Since later approximations replay previous samples they also include previously validated edges.
     }
\end{figure*}

\section{Effort Informed Roadmaps (\EI{})}

\EI{} extends EIT* to the multiquery setting.
EIT* is an almost-surely asymptotically optimal anytime sampling-based path planning algorithm that is based on an asymmetric search which simultaneously calculates and exploits problem-specific heuristics.
Both EIT* and \EI{} sample batches of states, and view these states as a series of edge-implicit random geometric graphs (RGGs) \cite{penrose2003random}, as in BIT* \cite{gammell_ijrr20}.
The edges in each RGG are processed in a reverse search informed by an \textit{a priori} heuristic.
The reverse search is computationally inexpensive since collisions are checked at a lower resolution than in the forward search, i.e., \emph{sparsely} checked.
The reverse search computes approximation-specific heuristic estimates of the cost and effort to reach the goal, and provides a lower bound on the resolution-optimal solution in the current RGG approximation.
The forward search is guided by the cost and effort heuristics that are calculated in the reverse search. 
EIT* and \EI{} both compute a suboptimality bound by inflating the resolution-optimal solution cost and only consider edges that satisfy this suboptimality bound.
During the forward search, edges are fully collision checked.
For more details on EIT*, see~\cite{strub_dphil21}.

In the multiquery setting, the RGG likely contains validated edges from solving previous queries that would require zero validation effort to reuse in a solution.
\EI{} leverages these zero-effort edges and avoids unbounded graph growth by modifying EIT*'s batch sampling (\cref{ssec:graph}) and reverse search (\cref{ssec:rev}) while using the same forward search (\cref{ssec:fwd}).
The batch sampling is modified to rewind the approximation of each query to the initial batch of samples. 
This approximation is then improved by replaying the same batches of samples as in previous queries in order to reuse effort.

The reverse search of \EI{} differs from EIT* in that the search of each query is ordered by estimated validation effort until an initial solution is found.
After finding an initial solution, \EI{} computes an admissible cost heuristic in its reverse search, and the forward search is ordered by cost.
\EI{} is illustrated in~\cref{fig:illustration}, and algorithmic details are presented in \cref{alg:eiprm,alg:expand,alg:get_new_sample,alg:best_rev_edge_improve_sol,alg:pop_fwd_full},  with modifications compared to EIT* in orange.

\EI{} maintains the almost-sure asymptotic optimality and probabilistic completeness of EIT*.
Initially only considering the first batch of samples and later adding previous samples to improve the cost results in the same behaviour as EIT* as the number of samples goes to infinity and does not alter formal properties.
The full proof for almost-sure asymptotic optimality for EIT*, which implies probabilistic completeness, is presented in~\cite{strub_dphil21}.

\subsection{Notation}
We denote the search space as $X\subseteq\mathbb{R}^n$, with the subspace occupied by obstacles, $X_\text{obs}\subset X$, and the free space, $X_\text{free} = \text{closure}(X \setminus X_\text{obs})$.
The $i$-th query consists of the start state, $x_{\text{start},i}\in X_\text{free}$, and a set of goal states, $X_{\text{goal},i}\subset X_\text{free}$.
The states that make up our current RGG are stored in $X_\text{RGG}\subset X_\text{free}$.
A path, $\pi$, consists of a series of states through the free space, $\forall s \in [0,1],\,\, \pi(s) \in X_\text{free}$.
The cost, $c$, of a path is $c(\pi) \in \mathbb{R}_{\geq0}$.
A solution to the $i$-th query is a path that starts at the start state and ends at a goal state, $\pi(0)=x_{\text{start},i}$ and $\pi(1)\in X_{\text{goal},i}$.

\begin{figure}[H]
  \centering
\begin{minipage}[t]{0.49\textwidth}
\vspace{0pt}
\input{algorithms/main_alg}%
\end{minipage}\hspace{1mm}
\begin{minipage}[t]{0.49\textwidth}
\vspace{0pt}
\input{algorithms/expand}%
\vspace{2.6mm}
\input{algorithms/refine}%
\vspace{2.6mm}
\input{algorithms/can_improve_sol}%
\vspace{2.67mm}
\input{algorithms/pop_fwd}%
\end{minipage}
\end{figure}
{\vskip-6ex}

We denote admissible estimates with a hat,~$\widehat{\cdot}$, possibly inadmissible estimates with a bar,~$\bar{\cdot}$, 
and previously computed labels, i.e., real numbers attached to specific states, with square brackets,~$l[\cdot]$. 

Admissible and inadmissible cost estimates between two states, $x_i$ and $x_j$, are denoted $\hat{c}(x_i, x_j)$, and $\bar{c}(x_i, x_j)$, respectively.
We assume that the admissible cost estimate is lower than the inadmissible cost estimate, i.e., $\forall x_i, x_j\in X,\,\, \hat{c}(x_i, x_j) \leq \bar{c}(x_i, x_j)$.

We use $\hat{g}(x_\text{t}) = \hat{c}(x_\text{start}, x_\text{t})$ as shorthand for an admissible cost heuristic to come to the target state, $x_\text{t}$, from the start, $\hat{h}(x_\text{t}) = \min_{x_\text{g}\in X_\text{goal}}\hat{c}(x_\text{t}, x_\text{g})$ to denote an admissible cost heuristic to go from a target state to a goal state, and $\bar{h}(x_\text{t}) = \min_{x_\text{g}\in X_\text{goal}}\bar{c}(x_\text{t}, x_\text{g})$ to denote an inadmissible heuristic to go from a target state to a goal state.
An admissible estimate for the total cost for a path going through a state, $x_\text{t}$, is then given by $\hat{f}(x_\text{t}) = \hat{g}(x_\text{t}) + \hat{h}(x_\text{t})$.

The possibly inadmissible estimate of the planning effort between two states, $x_i$ and $x_j$, is $\effort(x_i, x_j)$.
The possibly inadmissible estimate for the effort to come to a target state, $x_\text{t}$, from the start state, $x_\text{start}$, is $\apeffort(x_\text{t})$, which may be defined independently of $\effort$.

For two sets, $A$ and $B$, $A\setInsert B$ denotes $A\gets A \cup B$ and $A\setRemove B$ denotes $A\gets A\setminus B$.
The cardinality of a set is denoted by $|\cdot|$.

\subsection{Approximation}\label{ssec:graph}
\EI{} incrementally adds batches of $m$ states to the RGG to build a discrete approximation of the search space (\cref{alg:eiprm}, \cref{alg:add_samples}).
Informed sampling \cite{gammell2014informed} can be used to focus the approximation on the part of the space that can improve the solution once an initial solution to the current query has been found, if appropriate.

\EI{} considers connections between each sample and its $k$-nearest neighbours or states within a distance $r$ as well as previously validated edges independent of their distance to the state currently under consideration (\cref{alg:expand}, \cref{alg:add_valid_neighbours}).

If not handled explicitly, the size of the RGG will grow unbounded over the course of multiple queries.
\EI{} manages the growth of the graph by rewinding the sampling-based approximation to the first batch to find the initial solution to each query and by pruning starts and goals from the graph to limit graph growth.

\subsubsection{Batch Rewinding}
High-resolution approximations often contain high-quality solutions, but are computationally expensive to search due to the computational cost of the nearest-neighbour lookup and the required depth of the search.
When fast solution times are desired, low-resolution approximations are often better since the computational cost of these operations reduce with the number of samples.

\EI{} stores all sampled states $x_i$ for the duration of the multiquery problem in a buffer, $\mathcal{B}=(x_1, x_2, \cdots, x_n)$, and a new batch of samples is added to the approximation from the buffer when \texttt{refine\_approximation} is called (\cref{alg:get_new_sample}).
If the buffer does not contain enough samples, new states are first sampled and added to the buffer (\cref{alg:get_new_sample}, \cref{alg:sample_uni}).
Samples from the buffer are only added to the batch if they can improve the current solution (\cref{alg:get_new_sample}, \cref{alg:rejection}).
The current position in the buffer, $i_\text{buffer}$, is incremented as samples from the buffer are used (\cref{alg:get_new_sample}, \cref{alg:increment}), and is reset once a new planning query is considered (\cref{alg:eiprm}, \cref{alg:clear}).

\subsubsection{Start/Goal Pruning}
The size of the graph will grow unbounded with the number of queries if all starts and goals of every query are kept in the graph.
Keeping all starts and goals in the graph may also result in a nonuniform distribution of states if the starts and goals are not uniformly distributed.

Forgetting all starts and goals prevents unbounded graph growth and maintains the state sampling distribution, but discards the effort spent validating associated edges.
These conflicting behaviours are balanced by keeping the starts or goals in a buffer, $\mathcal{B}_\text{start/goal}$, if they satisfy a user-specified criterion (\cref{alg:eiprm}, \cref{alg:add_starts_to_buffer}).
The stored starts and goals are added to the RGG at the same time as the new start and goals of the current query (\cref{alg:eiprm}, \cref{alg:add_start_goals}) and the first batch of samples.\looseness=-1

\subsection{Reverse Search}\label{ssec:rev}

\EI{} first tries to find a solution as quickly as possible, and then tries to decrease the cost as quickly as possible.
The reverse search is therefore initially ordered on validation effort, and afterwards on cost.

The reverse search considers the best edge, starting at the source state, $x_\text{s}$, to the target state, $x_\text{t}$, from the edge-queue, $\mathcal{Q}_\mathcal{R}$.
This queue is lexicographically ordered by effort if no solution is available, 
\begin{equation}
    k^\text{effort}_{\mathcal{R}}(x_\text{s}, x_\text{t}) = \left(\effort[x_\text{s}] + \effort(x_\text{s}, x_\text{t}) + \apeffort(x_\text{t}), \hat{h}[x_\text{s}] + \hat{c}(x_\text{s}, x_\text{t}) + \hat{g}(x_\text{t})\right),
\end{equation}
and by cost once a solution was found,
\begin{equation}
    k^\text{cost}_{\mathcal{R}}(x_\text{s}, x_\text{t}) = \left(\hat{h}[x_\text{s}] + \hat{c}(x_\text{s}, x_\text{t}) + \hat{g}(x_\text{t}), \effort[x_\text{s}] + \effort(x_\text{s}, x_\text{t}) + \apeffort(x_\text{t})\right),
\end{equation}
where $\apeffort$ and $\hat{g}$ are \textit{a priori} estimates of the effort and cost to go, respectively.
The elements of the key are estimates of the total computational effort and the total solution cost of a path through an edge.
Their ordering depends on whether the current query already has a solution.

The edge with the lowest key is extracted (\cref{alg:eiprm}, \cref{alg:if_not_fin,alg:k_eiprm,alg:else_not_fin,alg:k_eit}) and collisions are checked sparsely at evenly distributed states along the edge (\cref{alg:eiprm}, \cref{alg:no_sparse_coll}).
If no collision is found, the computed cost heuristics, $\bar{h}[\cdot]$ and $\hat{h}[\cdot]$, and the effort heuristic, $\effort[\cdot]$, of the target state, $x_\text{t}$, are updated (\cref{alg:eiprm}, \cref{alg:update_inadmissible,alg:update_heuristic,alg:update_heuristic_2}).
The target state is then expanded, and the edges to its neighbours are inserted into the reverse queue, (\cref{alg:eiprm}, \cref{alg:expand_rev}) and the iteration restarts.
If a collision was found, the edge is added to the set of invalid edges, $E_\text{invalid}$ (\cref{alg:eiprm}, \cref{alg:add_sparse_invalid}).
The reverse search terminates when it is guaranteed to have found the resolution-optimal solution on the current RGG approximation or no solution is found, as in A* (\cref{alg:eiprm}, \cref{alg:can_improv_rev}), where $k^\text{effort}_{\mathcal{F}}$ and $k^\text{cost}_{\mathcal{F}}$ are the lexicographical sortings of the forward search and are defined analogously to the respective reverse keys.

\subsection{Forward Search} \label{ssec:fwd}
The forward search is based on anytime explicit estimation search (AEES) \cite{thayer2012better} and is guided by the calculated heuristics to effectively find solutions to each query.
This search completely checks edges for collision and is more computationally expensive than the reverse search.

The previously computed admissible cost heuristic provides a lower bound, $\hat{s}$, on the resolution-optimal solution in the current RGG,
\begin{equation}
    \hat{s} = \min_{(x_\text{s}, x_\text{t})\in\mathcal{Q}_\mathcal{F}} \left\{{g}_\mathcal{F}(x_\text{s}) + \hat{c}(x_\text{s}, x_\text{t}) + \hat{h}[x_\text{t}]\right\},
\end{equation}
where ${g}_\mathcal{F}(x_\text{s})$ is the cost to come through the forward tree to the source state, $x_\text{s}$, and $Q_\mathcal{F}$ is the edge-queue of the forward search.
A potentially more accurate estimate of the resolution-optimal cost can be calculated with the inadmissible cost heuristic,
\begin{equation}
    \bar{s} = \min_{(x_\text{s}, x_\text{t})\in\mathcal{Q}_\mathcal{F}} \left\{{g}_\mathcal{F}(x_\text{s}) + \bar{c}(x_\text{s}, x_\text{t}) + \bar{h}[x_\text{t}]\right\}.
\end{equation}
This possibly inadmissible estimate can be more accurate than its admissible counterpart since the inadmissible cost heuristic can use information that may overestimate the true cost.
The focal set, $\mathcal{S}$, is the set of edges that can possibly lead to a solution within the current suboptimality bound, $w\bar{s}$,
\begin{equation}
    \mathcal{S} = \left\{(x_\text{s}, x_\text{t}) \; |\;  {g}_\mathcal{F}(x_\text{t}) + \bar{c}(x_\text{t}, x_\text{s}) + \bar{h}[x_\text{s}] \leq w\bar{s}\right\}.
\end{equation}

\EI{} expands the next edge (\cref{alg:eiprm}, \cref{alg:pop_fwd}, \cref{alg:pop_fwd_full}) considering the focal set and fully collision checks the edge (\cref{alg:eiprm}, \cref{alg:coll_free}).
If the edge is found to be invalid, it is labeled as such (\cref{alg:eiprm}, \cref{alg:add_invalid}) and the reverse search is restarted.%

Edges are selected for expansion by first considering the minimum remaining validation effort in the focal set,
\begin{equation}
    \argmin_{(x_\text{s}, x_\text{t})\in \mathcal{S}} \left\{\effort(x_\text{s}, x_\text{t}) + \effort[x_\text{t}]\right\}.
\end{equation}
If this edge can improve the solution, it is selected (\cref{alg:pop_fwd_full}, \cref{alg:ret_min_effort}).
If not, the edge with the lowest inadmissible cost estimate is selected if it is estimated to lead to a solution within the current suboptimality bound (\cref{alg:pop_fwd_full}, \cref{alg:ret_min_inad_cost}).
Otherwise, the edge with the lowest admissible cost estimate is selected (\cref{alg:pop_fwd_full}, \cref{alg:ret_ad_cost}).

The forward search continues until it is known that the best edge in the forward queue can not improve the solution (\cref{alg:eiprm}, \cref{alg:can_improv_fwd}), or a solution is found.
If a solution is found, the best achieved cost is updated (\cref{alg:eiprm}, \cref{alg:1}), the search is then ordered by cost by setting the suboptimality factor to one (\cref{alg:eiprm}, \cref{alg:2}), the approximation is refined (\cref{alg:eiprm}, \cref{alg:add_samples}), and the loop restarts with the reverse search.

This search continues as long as time allows and almost-surely converges asymptotically to the optimal solution.

\section{Experiments \& Results}

We evaluated \EI{} on a set of simulated scenarios\footnote{All experiments were run using OMPL 1.5, on a laptop with an Intel i7-4720HQ CPU @ 2.60GHz processor with 16GB RAM.}, and compared it to a selection of both single-{} and multiquery planners available in OMPL: PRM*, LazyPRM*, RRT-Connect, RRT*, and EIT*.
The OMPL version of SPARS/SPARS2 was not included due to performance.

RRT* used a goal bias of 0.05.
Both RRT-based planners used maximum edge lengths of 0.3, 0.5, 1.25, and 2.4 in $\mathbb{R}^2$, $\mathbb{R}^4$, $\mathbb{R}^8$, and $\mathbb{R}^{14}$, respectively.
\EI{} and EIT* used the $k$-nearest neighbour method and sampled $100$ states per batch.
The \textit{a priori} heuristic for both admissible and inadmissible cost in \EI{} and EIT* was the Euclidean distance.
The \textit{a priori} heuristic for inadmissible effort between two states, i.e., $\effort$, in EIT* and \EI{} was the Euclidean distance divided by the needed remaining collision checking resolution.
In order to fully exploit the possibility of preexisting zero-effort edges, \EI{} used the zero heuristic for the inadmissible effort to come, i.e., $\apeffort$.\looseness=-1

In order to limit the growth of the graph, \EI{} kept starts and goals after a query if the number of required collision checks, i.e., the validation effort, to reach the state from the closest existing neighbour was larger than \num[group-separator={,}]{50000} and otherwise forgot them.

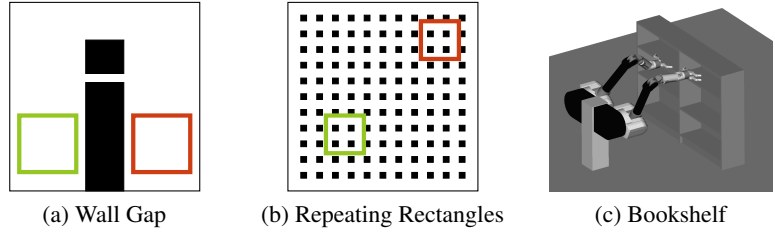
\begin{figure}[t]
\centering
    \begin{subfigure}[t]{.3\textwidth}
        \centering
        \begin{tikzpicture}[scale = 2.5] %
\draw [obstacle] (-0.1, -0.5) rectangle(0.1, 0.3);

\draw [boundary] (-0.5, -0.5) rectangle (0.5, 0.5);

\draw [fill = white, draw = none] (-0.11, 0.08) rectangle (0.11, 0.12);

\draw [start_region] (-0.45, -0.4) rectangle (-0.15, -0.1);

\draw [goal_region] (0.15, -0.4) rectangle (0.45, -0.1);
\end{tikzpicture}%
        
        \caption{\label{fig:wall_gap}Wall Gap}
    \end{subfigure}%
    \begin{subfigure}[t]{.3\textwidth}
        \centering
        \begin{tikzpicture}[scale = 2.5] %
\draw [obstacle] (-0.432, -0.432) rectangle  (-0.402, -0.402);
\draw [obstacle] (-0.348, -0.432) rectangle (-0.318, -0.402);
\draw [obstacle] (-0.265, -0.432) rectangle (-0.235, -0.402);
\draw [obstacle] (-0.182, -0.432) rectangle (-0.152, -0.402);
\draw [obstacle] (-0.0983, -0.432) rectangle (-0.0683, -0.402);
\draw [obstacle] (-0.015, -0.432) rectangle (0.015, -0.402);
\draw [obstacle] (0.0683, -0.432) rectangle (0.0983, -0.402);
\draw [obstacle] (0.152, -0.432) rectangle (0.182, -0.402);
\draw [obstacle] (0.235, -0.432) rectangle (0.265, -0.402);
\draw [obstacle] (0.318, -0.432) rectangle (0.348, -0.402);
\draw [obstacle] (0.402, -0.432) rectangle (0.432, -0.402);
\draw [obstacle] (-0.432, -0.348) rectangle (-0.402, -0.318);
\draw [obstacle] (-0.348, -0.348) rectangle (-0.318, -0.318);
\draw [obstacle] (-0.265, -0.348) rectangle (-0.235, -0.318);
\draw [obstacle] (-0.182, -0.348) rectangle (-0.152, -0.318);
\draw [obstacle] (-0.0983, -0.348) rectangle (-0.0683, -0.318);
\draw [obstacle] (-0.015, -0.348) rectangle (0.015, -0.318);
\draw [obstacle] (0.0683, -0.348) rectangle (0.0983, -0.318);
\draw [obstacle] (0.152, -0.348) rectangle (0.182, -0.318);
\draw [obstacle] (0.235, -0.348) rectangle (0.265, -0.318);
\draw [obstacle] (0.318, -0.348) rectangle (0.348, -0.318);
\draw [obstacle] (0.402, -0.348) rectangle (0.432, -0.318);
\draw [obstacle] (-0.432, -0.265) rectangle (-0.402, -0.235);
\draw [obstacle] (-0.348, -0.265) rectangle (-0.318, -0.235);
\draw [obstacle] (-0.265, -0.265) rectangle (-0.235, -0.235);
\draw [obstacle] (-0.182, -0.265) rectangle (-0.152, -0.235);
\draw [obstacle] (-0.0983, -0.265) rectangle (-0.0683, -0.235);
\draw [obstacle] (-0.015, -0.265) rectangle (0.015, -0.235);
\draw [obstacle] (0.0683, -0.265) rectangle (0.0983, -0.235);
\draw [obstacle] (0.152, -0.265) rectangle (0.182, -0.235);
\draw [obstacle] (0.235, -0.265) rectangle (0.265, -0.235);
\draw [obstacle] (0.318, -0.265) rectangle (0.348, -0.235);
\draw [obstacle] (0.402, -0.265) rectangle (0.432, -0.235);
\draw [obstacle] (-0.432, -0.182) rectangle (-0.402, -0.152);
\draw [obstacle] (-0.348, -0.182) rectangle (-0.318, -0.152);
\draw [obstacle] (-0.265, -0.182) rectangle (-0.235, -0.152);
\draw [obstacle] (-0.182, -0.182) rectangle (-0.152, -0.152);
\draw [obstacle] (-0.0983, -0.182) rectangle (-0.0683, -0.152);
\draw [obstacle] (-0.015, -0.182) rectangle (0.015, -0.152);
\draw [obstacle] (0.0683, -0.182) rectangle (0.0983, -0.152);
\draw [obstacle] (0.152, -0.182) rectangle (0.182, -0.152);
\draw [obstacle] (0.235, -0.182) rectangle (0.265, -0.152);
\draw [obstacle] (0.318, -0.182) rectangle (0.348, -0.152);
\draw [obstacle] (0.402, -0.182) rectangle (0.432, -0.152);
\draw [obstacle] (-0.432, -0.0983) rectangle (-0.402, -0.0683);
\draw [obstacle] (-0.348, -0.0983) rectangle (-0.318, -0.0683);
\draw [obstacle] (-0.265, -0.0983) rectangle (-0.235, -0.0683);
\draw [obstacle] (-0.182, -0.0983) rectangle (-0.152, -0.0683);
\draw [obstacle] (-0.0983, -0.0983) rectangle (-0.0683, -0.0683);
\draw [obstacle] (-0.015, -0.0983) rectangle (0.015, -0.0683);
\draw [obstacle] (0.0683, -0.0983) rectangle (0.0983, -0.0683);
\draw [obstacle] (0.152, -0.0983) rectangle (0.182, -0.0683);
\draw [obstacle] (0.235, -0.0983) rectangle (0.265, -0.0683);
\draw [obstacle] (0.318, -0.0983) rectangle (0.348, -0.0683);
\draw [obstacle] (0.402, -0.0983) rectangle (0.432, -0.0683);
\draw [obstacle] (-0.432, -0.015) rectangle (-0.402, 0.015);
\draw [obstacle] (-0.348, -0.015) rectangle (-0.318, 0.015);
\draw [obstacle] (-0.265, -0.015) rectangle (-0.235, 0.015);
\draw [obstacle] (-0.182, -0.015) rectangle (-0.152, 0.015);
\draw [obstacle] (-0.0983, -0.015) rectangle (-0.0683, 0.015);
\draw [obstacle] (-0.015, -0.015) rectangle (0.015, 0.015);
\draw [obstacle] (0.0683, -0.015) rectangle (0.0983, 0.015);
\draw [obstacle] (0.152, -0.015) rectangle (0.182, 0.015);
\draw [obstacle] (0.235, -0.015) rectangle (0.265, 0.015);
\draw [obstacle] (0.318, -0.015) rectangle (0.348, 0.015);
\draw [obstacle] (0.402, -0.015) rectangle (0.432, 0.015);
\draw [obstacle] (-0.432, 0.0683) rectangle (-0.402, 0.0983);
\draw [obstacle] (-0.348, 0.0683) rectangle (-0.318, 0.0983);
\draw [obstacle] (-0.265, 0.0683) rectangle (-0.235, 0.0983);
\draw [obstacle] (-0.182, 0.0683) rectangle (-0.152, 0.0983);
\draw [obstacle] (-0.0983, 0.0683) rectangle (-0.0683, 0.0983);
\draw [obstacle] (-0.015, 0.0683) rectangle (0.015, 0.0983);
\draw [obstacle] (0.0683, 0.0683) rectangle (0.0983, 0.0983);
\draw [obstacle] (0.152, 0.0683) rectangle (0.182, 0.0983);
\draw [obstacle] (0.235, 0.0683) rectangle (0.265, 0.0983);
\draw [obstacle] (0.318, 0.0683) rectangle (0.348, 0.0983);
\draw [obstacle] (0.402, 0.0683) rectangle (0.432, 0.0983);
\draw [obstacle] (-0.432, 0.152) rectangle (-0.402, 0.182);
\draw [obstacle] (-0.348, 0.152) rectangle (-0.318, 0.182);
\draw [obstacle] (-0.265, 0.152) rectangle (-0.235, 0.182);
\draw [obstacle] (-0.182, 0.152) rectangle (-0.152, 0.182);
\draw [obstacle] (-0.0983, 0.152) rectangle (-0.0683, 0.182);
\draw [obstacle] (-0.015, 0.152) rectangle (0.015, 0.182);
\draw [obstacle] (0.0683, 0.152) rectangle (0.0983, 0.182);
\draw [obstacle] (0.152, 0.152) rectangle (0.182, 0.182);
\draw [obstacle] (0.235, 0.152) rectangle (0.265, 0.182);
\draw [obstacle] (0.318, 0.152) rectangle (0.348, 0.182);
\draw [obstacle] (0.402, 0.152) rectangle (0.432, 0.182);
\draw [obstacle] (-0.432, 0.235) rectangle (-0.402, 0.265);
\draw [obstacle] (-0.348, 0.235) rectangle (-0.318, 0.265);
\draw [obstacle] (-0.265, 0.235) rectangle (-0.235, 0.265);
\draw [obstacle] (-0.182, 0.235) rectangle (-0.152, 0.265);
\draw [obstacle] (-0.0983, 0.235) rectangle (-0.0683, 0.265);
\draw [obstacle] (-0.015, 0.235) rectangle (0.015, 0.265);
\draw [obstacle] (0.0683, 0.235) rectangle (0.0983, 0.265);
\draw [obstacle] (0.152, 0.235) rectangle (0.182, 0.265);
\draw [obstacle] (0.235, 0.235) rectangle (0.265, 0.265);
\draw [obstacle] (0.318, 0.235) rectangle (0.348, 0.265);
\draw [obstacle] (0.402, 0.235) rectangle (0.432, 0.265);
\draw [obstacle] (-0.432, 0.318) rectangle (-0.402, 0.348);
\draw [obstacle] (-0.348, 0.318) rectangle (-0.318, 0.348);
\draw [obstacle] (-0.265, 0.318) rectangle (-0.235, 0.348);
\draw [obstacle] (-0.182, 0.318) rectangle (-0.152, 0.348);
\draw [obstacle] (-0.0983, 0.318) rectangle (-0.0683, 0.348);
\draw [obstacle] (-0.015, 0.318) rectangle (0.015, 0.348);
\draw [obstacle] (0.0683, 0.318) rectangle (0.0983, 0.348);
\draw [obstacle] (0.152, 0.318) rectangle (0.182, 0.348);
\draw [obstacle] (0.235, 0.318) rectangle (0.265, 0.348);
\draw [obstacle] (0.318, 0.318) rectangle (0.348, 0.348);
\draw [obstacle] (0.402, 0.318) rectangle (0.432, 0.348);
\draw [obstacle] (-0.432, 0.402) rectangle (-0.402, 0.432);
\draw [obstacle] (-0.348, 0.402) rectangle (-0.318, 0.432);
\draw [obstacle] (-0.265, 0.402) rectangle (-0.235, 0.432);
\draw [obstacle] (-0.182, 0.402) rectangle (-0.152, 0.432);
\draw [obstacle] (-0.0983, 0.402) rectangle (-0.0683, 0.432);
\draw [obstacle] (-0.015, 0.402) rectangle (0.015, 0.432);
\draw [obstacle] (0.0683, 0.402) rectangle (0.0983, 0.432);
\draw [obstacle] (0.152, 0.402) rectangle (0.182, 0.432);
\draw [obstacle] (0.235, 0.402) rectangle (0.265, 0.432);
\draw [obstacle] (0.318, 0.402) rectangle (0.348, 0.432);
\draw [obstacle] (0.402, 0.402) rectangle (0.432, 0.432);

\draw [boundary] (-0.5, -0.5) rectangle (0.5, 0.5);

\draw [start_region] (-0.3, -0.3) rectangle (-0.1, -0.1);

\draw [goal_region] (0.2, 0.2) rectangle (0.4, 0.4);

\end{tikzpicture}%
        \caption{\label{fig:repeating_rectangles}Repeating Rectangles}
    \end{subfigure}%
     \begin{subfigure}[t]{.3\linewidth}
        \centering
        \includegraphics[width=.8\linewidth]{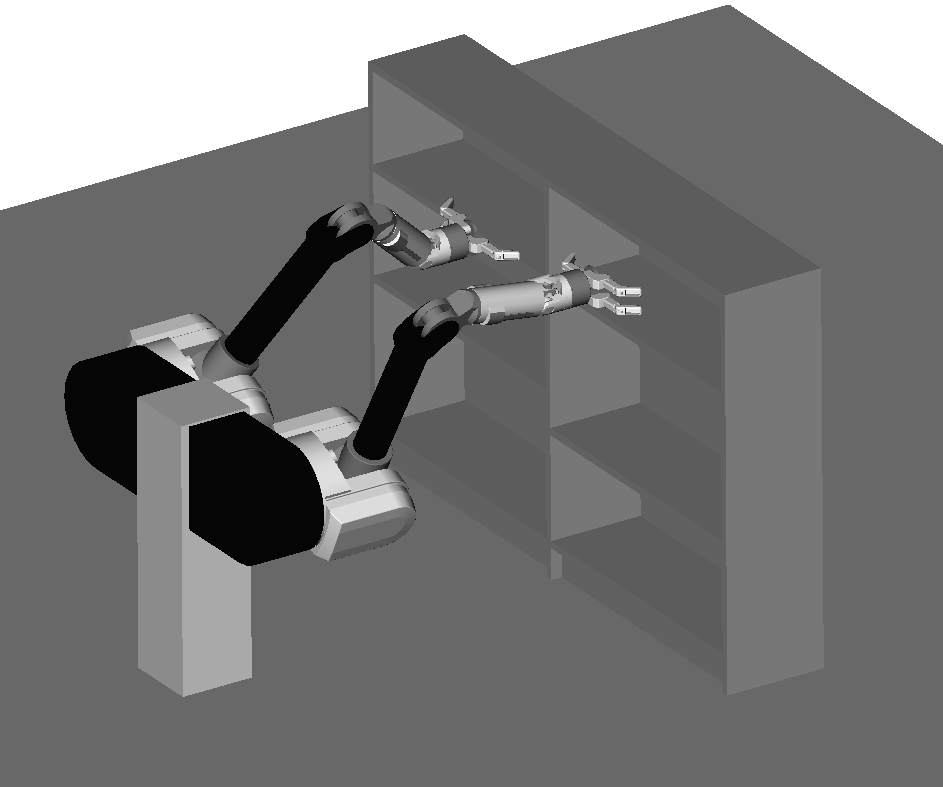} 
        \caption{\label{fig:bookshelf}Bookshelf}
    \end{subfigure} 
    
    \caption{\label{fig:abstract_scenarios} An illustration of the abstract scenarios in $\mathbb{R}^2$ with the subregions from which starts (green) and goals (red) are drawn uniformly, (a), (b). An illustration of the bookshelf scenario where the two-armed robot simulates picking/placing objects from the bookshelf, (c).
    Two versions of the repeating rectangles problem were considered, one in which start-goal queries are drawn from the subregions and the other in which they are drawn uniformly from the whole space.
    }
\end{figure}

\begin{table*}[t]
\vspace{-1em}
\caption{Cumulative median initial solution time, $t_\text{init}$, in seconds and cumulative median solution costs, $c_\text{init}$ and $c_\text{final}$, for all evaluated planners on a selection of scenarios.
    The cumulative median of a value is the median value of 100 runs summed over 100 queries.
    The initial cost is the cost of the first solution, while the final cost is the cost at the end of the planning time.
    Planners not run on a specific scenario are marked with -- and the \textbf{bold} value in each column is the best attained value.
    The $\infty$ indicates that it was not possible to compute the cumulative median for a planner due to the planner failing more than 50\% of the time on any query.
    Confidence intervals are not reported here due to space constraints. \Cref{fig:main_initial_duration,fig:cost_breakout} are representative of the relative confidence intervals for the respective experiments.
    }
\scriptsize
    \centering
    \sisetup{
        detect-weight=true,
        detect-inline-weight=math,
        table-format=3.2,
        group-minimum-digits=4,
        group-separator={,}
    }
    \renewcommand{\tabcolsep}{0.15pt}
    \begin{tabularx}{\textwidth}{
            X
            SSS 
            p{-5mm} 
            SSS 
            p{-5mm} 
            SSS 
            p{-5mm}
            SS[table-format=3.2]S}\toprule
        & \multicolumn{3}{c}{Wall Gap (\cref{fig:wall_gap})} & & \multicolumn{7}{c}{Repeating Rectangles (\cref{fig:repeating_rectangles})} & & \multicolumn{3}{c}{Bookshelf (\cref{fig:bookshelf})}  \\
        & \multicolumn{3}{c}{} & & \multicolumn{3}{c}{subregion} & & \multicolumn{3}{c}{global} & \multicolumn{3}{c}{} \\
        & \multicolumn{3}{c}{$\mathbb{R}^2$} & & \multicolumn{3}{c}{$\mathbb{R}^4$} & & \multicolumn{3}{c}{$\mathbb{R}^8$} & & \multicolumn{3}{c}{$\mathbb{R}^{14}$} \\
        {} & $t_\text{init}$ & $c_\text{init}$ & $c_\text{final}$ & & $t_\text{init}$ & $c_\text{init}$ & $c_\text{final}$ & & $t_\text{init}$ & $c_\text{init}$ & $c_\text{final}$ & & $t_\text{init}$ & $c_\text{init}$ & $c_\text{final}$ \\
        \midrule
PRM* & 4.21 & 98.9 & 98.1 & & 73.9 & 137.2 & 136.4 & & 18.6 & 206.4 & 204.8 & &  {--}  &  {--}  &  {--}  \\ 
LazyPRM* & 2.34 & \bfseries 95.4 & \bfseries 94.5 & & 25.7 & \bfseries 112.6 & \bfseries 111.7 & & 24.0 & \bfseries 166.3 & 165.0 & & 20.5 & \bfseries 520.0 & \bfseries 504.8 \\ 
RRT-Connect & 1.53 & 185.1 & 185.1 & & 3.18 & 163.5 & 163.5 & & 4.55 & 257.7 & 257.7 & & 45.9 & 884.2 & 884.2 \\ 
RRT* & 5.67 & 162.1 & 121.4 & & 24.1 & 155.5 & 134.2 & & 23.8 & 269.3 & 233.0 & &  {--}  &  {--}  &  {--}  \\ 
EIT* & 1.51 & 145.9 & 95.9 & & 3.00 & 157.9 & 112.1 & & 3.21 & 200.8 & \bfseries 147.3 & & $\infty$ & $\infty$ & $\infty$ \\ 
\EI & \bfseries 0.28 & 153.4 & 96.5 & & \bfseries 0.77 & 195.6 & 113.7 & & \bfseries 2.0 & 414.4 & 155.7 & & \bfseries 10.4 & 987.1 & 509.4 \\ 

        \bottomrule
    \end{tabularx}
    \label{tab:all_summary}
\end{table*}

\begin{figure}[t]
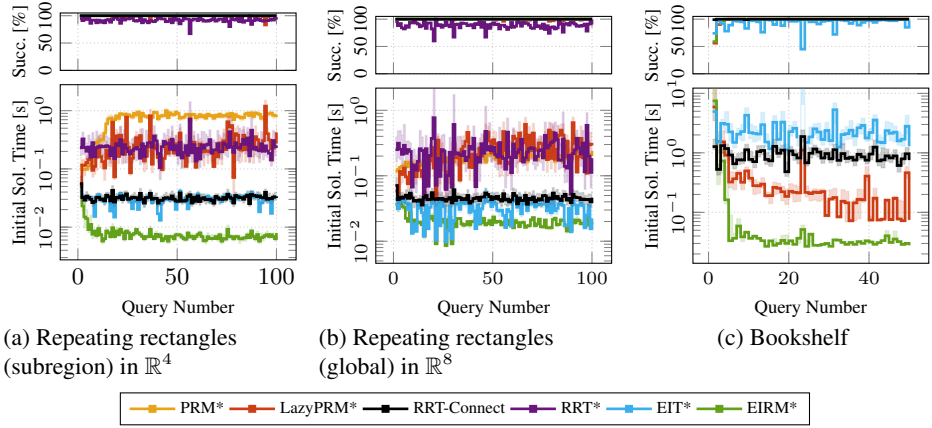

\captionsetup[subfigure]{aboveskip=0pt}
    \centering
    \begin{subfigure}[t]{.32\textwidth}
        \centering
        \input{img/tikz/repeated_rectangles/4d_sub_isrr.tikz}
        \caption{Repeating rectangles\\ (subregion) in $\mathbb{R}^4$}
    \end{subfigure}\hfill%
    \begin{subfigure}[t]{.32\textwidth}
        \centering
        \input{img/tikz/repeated_rectangles/8d_uni_isrr.tikz}
        \caption{Repeating rectangles\\ (global) in $\mathbb{R}^8$}
    \end{subfigure}\hfill%
    \begin{subfigure}[t]{0.32\textwidth}
        \centering
        \input{img/tikz/robots/bookshelf_isrr_2.tikz}
        \caption{\label{fig:bookshelf_initial_sol}Bookshelf}
    \end{subfigure}%
    \\[0.5em]
    \begin{subfigure}[b]{1.0\textwidth}%
        \centering
        \begin{tikzpicture}
\begin{axis} [
  width=\textwidth,
  height=0.5\textwidth,
  unbounded coords=jump,
  xtick align=inside,
  ytick align=inside,
  anchor=north,
  hide axis,
  xmajorgrids,
  ymajorgrids,
  major grid style={densely dotted, black!20},
  xmin=0,
  xmax=10,
  ymin=0,
  ymax=10,
  xlabel style={font=\footnotesize},
  xticklabel style={font=\footnotesize},
  ylabel style={font=\footnotesize},
  yticklabel style={font=\footnotesize},
  legend style={anchor=south, legend cell align=left, legend columns=-1, at={(axis cs:5, 6)}, nodes={scale=0.7, transform shape}}
]
\addlegendimage{espyellow, line width = 1.0pt, mark size=1.0pt, mark=square*}
\addlegendentry{PRM*}
\addlegendimage{esplightred, line width = 1.0pt, mark size=1.0pt, mark=square*}
\addlegendentry{LazyPRM*}
\addlegendimage{espblack, line width = 1.0pt, mark size=1.0pt, mark=square*}
\addlegendentry{RRT-Connect}
\addlegendimage{esppurple, line width = 1.0pt, mark size=1.0pt, mark=square*}
\addlegendentry{RRT*}
\addlegendimage{esplightblue, line width = 1.0pt, mark size=1.0pt, mark=square*}
\addlegendentry{EIT*}
\addlegendimage{espgreen, line width = 1.0pt, mark size=1.0pt, mark=square*}
\addlegendentry{\EI{}}

\end{axis}
\end{tikzpicture}
    \end{subfigure}
    \caption{\label{fig:main_initial_duration} 
    The median initial solution times per query over 100 runs for the repeating rectangles with subregion starts and goals in $\mathbb{R}^4$, (a), globally sampled starts and goals in $\mathbb{R}^8$, (b), and the bookshelf scenario, (c).
    The solid line is the median initial solution time per query, and the shaded area is the nonparametric 99\% confidence interval.
    Unsuccessful runs are treated as having infinite cost.
    For the bookshelf scenario, PRM* and RRT* were not run due to performance.\looseness=-1
    }
\end{figure}

\begin{figure}[t]
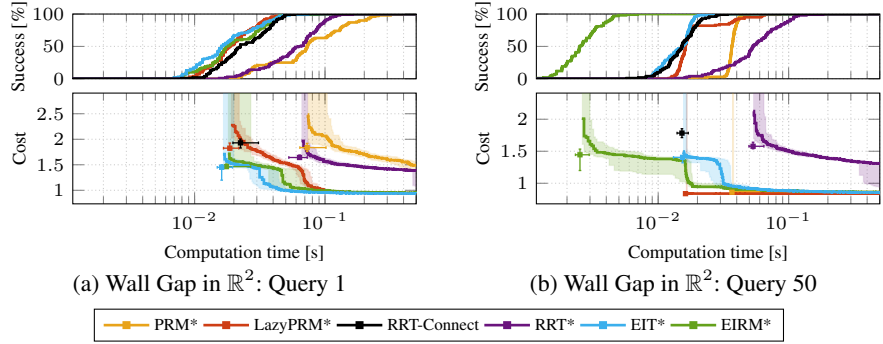

\centering
\captionsetup[subfigure]{aboveskip=0pt}
    \begin{subfigure}[t]{.5\textwidth}
        \centering
        \input{img/tikz/wall_gap/cost_breakout/q0.tikz}
        \caption{Wall Gap in $\mathbb{R}^2$: Query 1}
    \end{subfigure}%
    \begin{subfigure}[t]{.5\textwidth}
        \centering
        \input{img/tikz/wall_gap/cost_breakout/q50.tikz}
        \caption{Wall Gap in $\mathbb{R}^2$: Query 50}
    \end{subfigure}
    \\[0.5em]
    \begin{subfigure}[b]{1.0\textwidth}%
        \centering
        \begin{tikzpicture}
\begin{axis} [
  width=\textwidth,
  height=0.5\textwidth,
  unbounded coords=jump,
  xtick align=inside,
  ytick align=inside,
  anchor=north,
  hide axis,
  xmajorgrids,
  ymajorgrids,
  major grid style={densely dotted, black!20},
  xmin=0,
  xmax=10,
  ymin=0,
  ymax=10,
  xlabel style={font=\footnotesize},
  xticklabel style={font=\footnotesize},
  ylabel style={font=\footnotesize},
  yticklabel style={font=\footnotesize},
  legend style={anchor=south, legend cell align=left, legend columns=-1, at={(axis cs:5, 6)}, nodes={scale=0.7, transform shape}}
]
\addlegendimage{espyellow, line width = 1.0pt, mark size=1.0pt, mark=square*}
\addlegendentry{PRM*}
\addlegendimage{esplightred, line width = 1.0pt, mark size=1.0pt, mark=square*}
\addlegendentry{LazyPRM*}
\addlegendimage{espblack, line width = 1.0pt, mark size=1.0pt, mark=square*}
\addlegendentry{RRT-Connect}
\addlegendimage{esppurple, line width = 1.0pt, mark size=1.0pt, mark=square*}
\addlegendentry{RRT*}
\addlegendimage{esplightblue, line width = 1.0pt, mark size=1.0pt, mark=square*}
\addlegendentry{EIT*}
\addlegendimage{espgreen, line width = 1.0pt, mark size=1.0pt, mark=square*}
\addlegendentry{\EI{}}

\end{axis}
\end{tikzpicture}
    \end{subfigure}
    \caption{\label{fig:cost_breakout}
    The planner performance in two queries for the wall gap in $\mathbb{R}^2$.
    The success plots (top) show the percentage of successful runs over time.
    The cost evolution plots (bottom) show the median cost at a given time as a thick line, with the nonparametric 99\% confidence interval as shaded area.
    The squares show the median initial solution time for the query and the corresponding median initial cost.
    The absence of a solution is treated as having an infinite cost.
    }
\end{figure}

\subsection{Abstract Scenarios}\label{ssec:abstract_exp}
We considered two abstract scenarios with different obstacle configurations in $\mathbb{R}^2$, $\mathbb{R}^4$, and $\mathbb{R}^8$ (\cref{fig:abstract_scenarios}).
The scenario in \cref{fig:repeating_rectangles} was tested with the starts and goals were sampled uniformly at random from both subregions and sampled uniformly at random over the whole search space.
The subregion scenario often occurs in construction or warehouse settings where robots move between two regions.

Each planner was run $100$ times with different pseudorandom seeds on a multiquery problem consisting of a sequence of $100$ different queries. 
The query sequence was defined for each problem by randomly sampling $100$ starts and goals, and the same random sequence was used for all $100$ runs of all planners.
The maximum runtime per query was $0.5\text{s}$, $2\text{s}$, and $2\text{s}$ in $\mathbb{R}^2$, $\mathbb{R}^4$, and $\mathbb{R}^8$, respectively.
The collision detection resolution was set to $5\cdot 10^{-6}$ in the abstract problems to imitate the computational cost of collision checking of the robotic experiment, as in \cite{strub_dphil21}.

The median initial solution time per query along with confidence intervals for the repeating rectangles with subregion starts and goals in $\mathbb{R}^4$, and globally sampled starts and goals in $\mathbb{R}^8$ are shown in \cref{fig:main_initial_duration}. 
\Cref{tab:all_summary} summarizes both the cumulative median initial solution time across all queries (i.e., the integral of the plots shown in~\cref{fig:main_initial_duration}), and the corresponding cumulative median initial cost along with the cumulative median final cost of all planners.
The evolution of the cost for the 1$^\text{st}$ and the 50$^\text{th}$ query on the example of the wall gap in $\mathbb{R}^2$ is presented in \cref{fig:cost_breakout}.

The initial solution time achieved by \EI{} is faster than the time achieved by all the other planners.
The initial cost for the subregion scenarios is comparable to RRT-Connect, while in the globally sampled setting the cost is higher than the cost of the other planners since they use more computational time to find a initial solution.
In both the subregion and the globally sampled start-goal scenarios, \EI{} converges to a solution that is similar to the other optimizing planners when given the same amount of computational time.

\subsection{Robotic Scenario}
We considered a two-armed robot ($\mathbb{R}^{14}$) with the queries chosen such that they simulate rearranging objects on a bookshelf (\cref{fig:bookshelf}).
Each planner was run $100$ times with different pseudorandom seeds on a sequence of $50$ different queries and was run for $10\text{s}$ for each query.
As for the abstract experiments, the sequence of random starts and goals was constant for all attempts.
The Flexible Collision Library (FCL) \cite{fcl} was used for collision checking, and the collision detection resolution was set to $0.036$ for the bookshelf scenario, as in \cite{strub_dphil21}. %

\Cref{fig:bookshelf_initial_sol} shows the initial solution time taken per query.
\Cref{tab:all_summary} again summarizes the cumulative initial solution time across all queries and the corresponding cumulative median initial and final costs.
\EI{} achieves up to an order-of-magnitude faster initial solutions for some queries and is approximately twice as fast cumulatively compared to the other planners.

\subsection{Initial Solutions} \label{app:init_sol}
The relative benefits of explicitly reusing previous search effort and managing graph size are evaluated by limiting planners to only finding an initial solution. 
Not letting the planners run until convergence reduces the problems of unbounded graph growth for PRM* and LazyPRM*.

These experiments were run for the repeating rectangles scenario with subregion start-goal queries in $\mathbb{R}^8$ and for the bookshelf experiment.
The experimental setup is the same as previously for both scenarios, but with early stopping after finding a solution.

\begin{figure}[t]
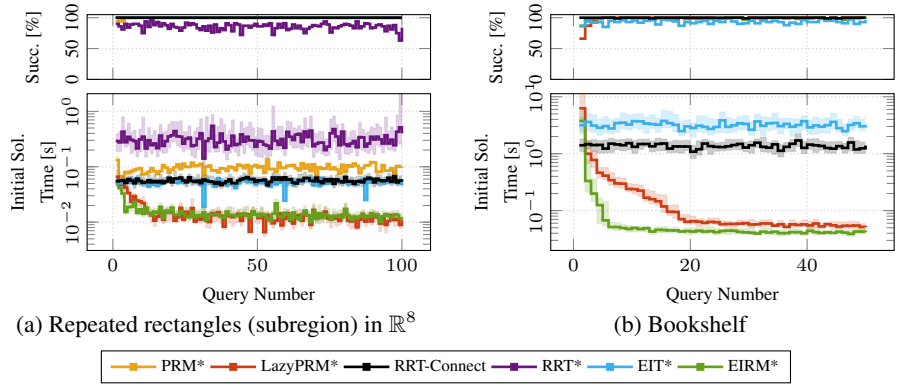

\centering
\captionsetup[subfigure]{aboveskip=0pt}
    \begin{subfigure}[t]{.5\textwidth}
        \centering
        \input{img/tikz/repeated_rectangles/initial/8d_sub.tikz}
        \caption{Repeated rectangles (subregion) in $\mathbb{R}^8$}
    \end{subfigure}%
    \begin{subfigure}[t]{.5\textwidth}
        \centering
        \input{img/tikz/robots/initial/bookshelf_initial.tikz}
        \caption{Bookshelf}
    \end{subfigure}
    \\[0.5em]
    \begin{subfigure}[b]{1.0\textwidth}%
        \centering
        \begin{tikzpicture}
\begin{axis} [
  width=\textwidth,
  height=0.5\textwidth,
  unbounded coords=jump,
  xtick align=inside,
  ytick align=inside,
  anchor=north,
  hide axis,
  xmajorgrids,
  ymajorgrids,
  major grid style={densely dotted, black!20},
  xmin=0,
  xmax=10,
  ymin=0,
  ymax=10,
  xlabel style={font=\footnotesize},
  xticklabel style={font=\footnotesize},
  ylabel style={font=\footnotesize},
  yticklabel style={font=\footnotesize},
  legend style={anchor=south, legend cell align=left, legend columns=-1, at={(axis cs:5, 6)}, nodes={scale=0.7, transform shape}}
]
\addlegendimage{espyellow, line width = 1.0pt, mark size=1.0pt, mark=square*}
\addlegendentry{PRM*}
\addlegendimage{esplightred, line width = 1.0pt, mark size=1.0pt, mark=square*}
\addlegendentry{LazyPRM*}
\addlegendimage{espblack, line width = 1.0pt, mark size=1.0pt, mark=square*}
\addlegendentry{RRT-Connect}
\addlegendimage{esppurple, line width = 1.0pt, mark size=1.0pt, mark=square*}
\addlegendentry{RRT*}
\addlegendimage{esplightblue, line width = 1.0pt, mark size=1.0pt, mark=square*}
\addlegendentry{EIT*}
\addlegendimage{espgreen, line width = 1.0pt, mark size=1.0pt, mark=square*}
\addlegendentry{\EI{}}

\end{axis}
\end{tikzpicture}
    \end{subfigure}
    \caption{\label{fig:appendix_non_converged_initial_sol} 
    The success plot (top) and the median initial solution times per query (bottom) for two experiments when the planning process was stopped as soon as an initial solution was found.
    In the success plot, the line indicates the number of runs that have solved the query at the end of the given planning time.
    In the median initial solution plot, the solid line is the median initial solution time per query, and the shaded area is the nonparametric 99\% confidence interval.
    Unsuccessful runs are treated as having infinite cost.
    For the bookshelf scenario, PRM* and RRT* were not run due to performance.
    }
\end{figure}

\Cref{fig:appendix_non_converged_initial_sol} shows the initial-solution time plots for the experiments.
\EI{} still achieves better median initial solution times in the bookshelf scenario and comparable times for the repeating rectangle scenario demonstrating the value of explicitly reusing information on more difficult problems.

\begin{figure}[t]
\captionsetup[subfigure]{aboveskip=-1pt}

    \centering
    \begin{subfigure}[t]{0.29\textwidth}
        \centering
        \vspace{-30mm}
        \includegraphics[width=.97\linewidth]{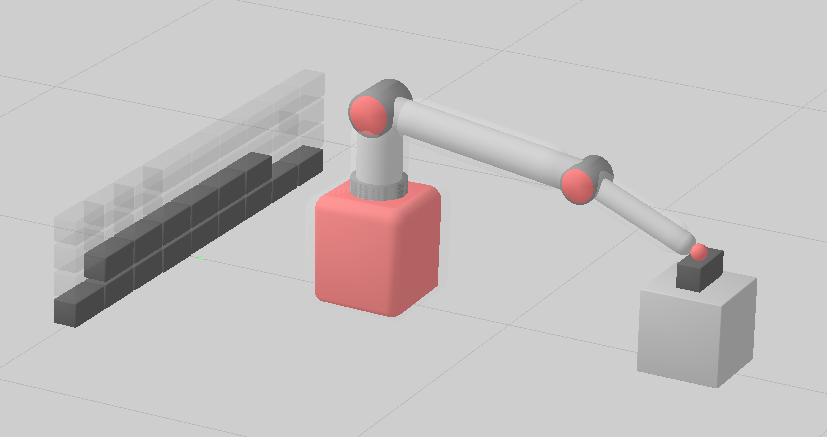}
        
        \vspace{3mm}
        \begin{tikzpicture}
\pgfplotsset{
compat=1.11,
legend image code/.code={
\draw[mark repeat=2,mark phase=2]
plot coordinates {
(0cm,0cm)
(0.15cm,0cm)        %
(0.3cm,0cm)         %
};%
}
}
\begin{axis} [
  width=\textwidth,
  height=0.5\textwidth,
  unbounded coords=jump,
  xtick align=inside,
  ytick align=inside,
  anchor=north,
  hide axis,
  xmajorgrids,
  ymajorgrids,
  major grid style={densely dotted, black!20},
  xmin=0,
  xmax=10,
  ymin=0,
  ymax=10,
  xlabel style={font=\scriptsize},
  xticklabel style={font=\scriptsize},
  ylabel style={font=\scriptsize},
  yticklabel style={font=\scriptsize},
  legend style={anchor=south, legend cell align=left, legend columns=-1, at={(axis cs:5, 6)}, nodes={scale=0.7, transform shape}}
]
\addlegendimage{espblack, line width = 1.0pt, mark size=1.0pt, mark=square*}
\addlegendentry{\footnotesize RRT-Connect}
\addlegendimage{esplightblue, line width = 1.0pt, mark size=1.0pt, mark=square*}
\addlegendentry{\footnotesize EIT*}
\addlegendimage{espgreen, line width = 1.0pt, mark size=1.0pt, mark=square*}
\addlegendentry{\footnotesize \EI{}}

\end{axis}
\end{tikzpicture}
        \caption{}

    \end{subfigure}%
    \begin{subfigure}[t]{.36\textwidth}
        \centering
        \input{img/tikz/robots/wall_stacking.tikz}
        \caption{}
    \end{subfigure}%
    \begin{subfigure}[t]{.3\textwidth}
        \centering
        \vspace{4.5mm}

\vspace{-3.2cm}
\scriptsize
    \centering
    \sisetup{
        detect-weight=true,
        detect-inline-weight=math,
        table-format=3.2,
        group-minimum-digits=4,
        group-separator={,}
    }
    \renewcommand{\tabcolsep}{0.2pt}
    \begin{tabularx}{\textwidth}{
            X
            SSS 
            }\toprule
        {} & $t_\text{init}$ & $c_\text{init}$ & $c_\text{final}$\\
        \midrule
RRT-Connect & 131.2 & \bfseries 221.5 & 221.5 \\ 
EIT* & 207.1 & 262.2 & 243.3 \\ 
\EI & \bfseries 76.3 & 288.7 & \bfseries 214.2\\ 

        \bottomrule
    \end{tabularx}
    \label{tab:walll}

        \vspace{9mm}
        \caption{}
    \end{subfigure}

    \caption{\label{fig:wall_stacking} 
    An illustration of the construction scenario showing the mobile manipulate picking up a brick with translucent bricks illustrating the target position of future bricks, (a). 
    The associated planner performance, (b), shows the success (top) and median initial solution times per query (bottom) over 25 runs and the culmulative median initial times and culmulative initial and final costs are shown in Table (c). 
    In the success plot, the line indicates the percentage of runs that have solved the query in the given planning time.
    In the median initial solution plot, the solid line is the median initial solution time per query and the shaded area is the nonparametric 99\% confidence interval.
    Unsuccessful runs are treated as having infinite cost.
    }
    \vspace{-0mm}
\end{figure}

\subsection{Construction Scenario}

Task and Motion Planning (TAMP) problems often pose multiquery scenarios where the environments changes and existing edges in the roadmap may be invalidated. 
We demonstrate a basic modification of \EI{} in a simplified construction setting \cite{21-hartmann-long} where a mobile manipulator ($\mathbb{R}^8$) stacks 36 bricks to build a wall~(\cref{fig:wall_stacking}).
The bricks that make up the wall are all the same, and are all picked up in the same location, simulating a conveyor belt that brings the bricks to the robot.

There are typically two planning problems in such a scenario.
The first is picking up the bricks and stacking it on the wall, the second is returning to the pickup location.
These two problems are often treated independently since the collision-checking envelope of the robot is different with and without a brick. 
We demonstrate the second scenario where the robot returns from placing a brick to pick up a new brick.

\EI{} was modified to remove the edges and vertices in its roadmap invalidated by newly placed bricks.
It was not possible to efficiently make such modifications to PRM* and LazyPRM*, so
\EI{} was only compared to the single-query planners which require no modifications, RRT-Connect and EIT*.
The planners were run $25$ times with different pseudorandom seeds on a $36$ query sequence with $10$s for each query.
FCL was used for collision checking.

The median initial time plot (\cref{fig:wall_stacking}) is promising.
While the time needed by RRT-Connect increases as the wall is built and the environment becomes more complex, the time taken by \EI{} decreases with the number of queries.
Future work will focus on fully adapting \EI{} to changing environments by developing more efficient ways to remove invalidated edges and vertices.
\section{Discussion}
\EI{} consistently outperforms all tested planners in the time necessary to find an initial solution.
The difference to other planners is most pronounced when queries are between subregions, since the previous paths are more likely to be part of future solutions.
An improvement of the initial query time can still be observed in problems with uniformly distributed starts and goals.
The cost of the quick initial solution from \EI{} is usually higher than the cost of the paths from other planners but the final cost is similar to the cost of other almost-surely asymptotically optimal planners.

\subsection{Initial Solution Time}
\Cref{fig:main_initial_duration} and \cref{tab:all_summary} show that \EI{} finds initial solutions up to an order-of-magnitude faster than the other tested planners.
It does this by explicitly seeking to reuse previous search effort and rewinding the approximation.

LazyPRM* fails to solve some of the tested problems reliably due to the growing graph size when improving solution cost.
In the experiments where the planners were stopped when an initial solution is found to limit this growth, \EI{} still achieves similar or better results than LazyPRM* on initial solution time.
\EI{} also needs fewer queries to benefit from previously invested effort compared to LazyPRM* since \EI{} explicitly tries to reuse validated edges.

Rewinding the approximation of the environment to the first batch of samples makes the performance of the planner independent of the previous query's final resolution since every query starts from a coarse resolution.
This may remove important paths that were found in later approximations, e.g., narrow passages.
Future work could investigate promoting \textit{promising} samples to earlier batches by ordering the samples in the buffer with an \emph{importance} metric.
This could lead to both quicker and higher quality initial solutions.
Similarly, pruning starts and goals too aggressively might lead to a loss of invested effort.
In future work, we intend to investigate the start and goal pruning method.

It might be beneficial to explore other heuristics for the possibly inadmissible effort, and the stopping conditions for the effort ordered reverse search.
We noticed actively reducing validation effort means that in some experiments validation effort is no longer the main computational cost of \EI{}.
This suggests that future work could include other time intensive steps of the algorithm in the effort heuristic, e.g., nearest neighbour lookups, which took up to 30\% of the planning time in our setting.
We currently run the reverse search until no solution candidate with a lower remaining validation effort exists. 
It might be faster overall to stop the reverse search earlier, and use an earlier solution candidate even if it may not be the path with the minimum remaining validation effort.

\subsection{Objective Value}
\Cref{tab:all_summary} reports the initial and final costs of the solution.
The initial path cost found by \EI{} is usually higher than LazyPRM*, but its final cost is within a few percentage points of the best found solution.
In some cases, \EI{} appeared not to converge efficiently to the best solution when the optimal solution was close to the straight-line path.
This may be due to rejection sampling from the sample buffer to obtain informed samples when refining the RGG.
Future work may consider how to efficiently sample the informed set while maintaining the uniform distribution of samples in the buffer.

If a suboptimal solution is acceptable, it might be desirable to smoothly interpolate between an effort-ordered and a cost-ordered search to allow for more path reuse.
This could be achieved with multi-objective A* \cite{stewart1991moa}.
The labels for the cost and effort would then not only depend on the state itself, but also on which path was taken to get to the state.
It is future work to investigate how to best incorporate this approach in \EI{}.

\vspace{-1mm}
\section{Conclusion}
Multiquery planners aim to efficiently solve multiple diverse motion planning problems in the same environment.
This is generally achieved by keeping the approximation built during the previous queries.
This can speed up the planning process, but few planners fully exploit the invested effort.

This paper presents \EI{}, a planner that explicitly aims to find paths with a low remaining validation effort.
This is achieved by using an asymmetric search that calculates cost and effort heuristics in a computationally cheap reverse search.
The heuristics are then used to guide the forward search, in which the edges are fully collision checked.\looseness=-1

\EI{} demonstrates that explicitly reusing computational effort and managing graph size between queries finds an initial solution quickly and then rapidly improves it.
This is shown to outperform existing state-of-the-art planners on initial solution time  while achieving similar solution quality on multiple different planning scenarios consisting of low-{} and high-dimensional abstract problems and robotic simulations.
Information on the OMPL implementation of \EI{} is available at {\urlstyle{same}\url{https://robotic-esp.com/code}}.

\vspace{-1mm}
{\footnotesize
\paragraph{Acknowledgement}
This research has been supported by the Deutsche Forschungsgemeinschaft (DFG, German Research Foundation) under Germany’s Excellence Strategy – EXC 2120/1 – 390831618 and UK Research and Innovation and EPSRC through ACE-OPS: From Autonomy to Cognitive assistance in Emergency OPerationS [EP/S030832/1]. }
\vspace{-2mm}

\renewcommand{\bibname}{References}
\let\oldbibliography\bibliography%
\renewcommand{\bibliography}[1]{{%
  \let\chapter\section%
  \oldbibliography{#1}}}%

\bibliographystyle{spbasic}
{
\footnotesize
\bibliography{short_ref}
}

\end{document}